\documentclass[12pt]{article}

\usepackage{geometry}
\usepackage{graphicx}
\usepackage{amsmath, amssymb}
\usepackage{booktabs}
\usepackage{multirow}
\usepackage{array}
\usepackage{xcolor}
\usepackage{hyperref}
\hypersetup{
	colorlinks=true,
	linkcolor=blue,
	citecolor=blue,
	urlcolor=blue,
}
\usepackage{natbib}  
\usepackage{adjustbox}
\usepackage{caption}
\usepackage{subcaption}
\usepackage{float}

\newcommand{\keywords}[1]{\vspace{0.5cm}\noindent\textbf{Keywords:} #1}

\usepackage{algorithm}
\usepackage{algpseudocode}

\begin{document}
	
	\title{A Hybrid Mamba-SAM Architecture for Efficient 3D Medical Image Segmentation}
	
	\author{
		\begin{tabular}{c}
			Mohammadreza Gholipour Shahraki\textsuperscript{1*} \\
			Mehdi Rezaeian\textsuperscript{2} \\
			Mohammad Ghasemzadeh\textsuperscript{2,*} \\
		\end{tabular}
	}
	\date{} 
	
	\maketitle
	
	\begin{center}
		\small
		\textsuperscript{1}Department of Electrical and Computer Engineering, Isfahan University of Technology, Isfahan, Iran \\
		\textsuperscript{2}Department of Computer Engineering, Yazd University, Yazd, Iran \\
		*Corresponding author: gholipour.m@ec.iut.ac.ir
	\end{center}
	
	\vspace{0.5cm}
	
	\begin{abstract}
		Accurate segmentation of 3D medical images such as MRI and CT is essential for clinical diagnosis and treatment planning. Foundation models like the Segment Anything Model (SAM) provide powerful general-purpose representations but struggle in medical imaging due to domain shift, their inherently 2D design, and the high computational cost of fine-tuning. To address these challenges, we propose Mamba-SAM, a novel and efficient hybrid architecture that combines a frozen SAM encoder with the linear-time efficiency and long-range modeling capabilities of Mamba-based State Space Models (SSMs). We investigate two parameter-efficient adaptation strategies. The first is a dual-branch architecture that explicitly fuses general features from a frozen SAM encoder with domain-specific representations learned by a trainable VMamba encoder using cross-attention. The second is an adapter-based approach that injects lightweight, 3D-aware Tri-Plane Mamba (TP-Mamba) modules into the frozen SAM ViT encoder to implicitly model volumetric context. Within this framework, we introduce Multi-Frequency Gated Convolution (MFGC), which enhances feature representation by jointly analyzing spatial and frequency-domain information via 3D discrete cosine transforms and adaptive gating. Extensive experiments on the ACDC cardiac MRI dataset demonstrate the effectiveness of the proposed methods. The dual-branch Mamba-SAM-Base model achieves a mean Dice score of 0.906, comparable to UNet++ (0.907), while outperforming all baselines on Myocardium (0.910) and Left Ventricle (0.971) segmentation. The adapter-based TP\_MFGC variant offers superior inference speed (4.77 FPS) with strong accuracy (0.880 Dice). These results show that hybridizing foundation models with efficient SSM-based architectures provides a practical and effective solution for 3D medical image segmentation.
	\end{abstract}
	
	
	\section{Introduction}
	The precise delineation of anatomical structures and pathological regions within medical images is a cornerstone of modern healthcare \cite{bernard2018}. Semantic segmentation, particularly for three-dimensional (3D) volumetric data acquired from modalities like Magnetic Resonance Imaging (MRI) and Computed Tomography (CT), provides crucial quantitative information for clinical diagnosis, surgical planning, radiation therapy targeting, and longitudinal monitoring of disease progression. For decades, the U-Net architecture \cite{ronneberger2015} and its numerous derivatives \cite{cciccek2016, oktay2018, isensee2021} have dominated the field, leveraging Convolutional Neural Networks (CNNs) to excel at capturing local image features and spatial hierarchies. However, the inherently limited receptive field of standard CNNs hinders their ability to effectively model long-range spatial dependencies and global contextual relationships, which are often vital for accurately interpreting complex anatomical scenes or identifying subtle, distributed pathologies.
	
	The advent of the Transformer architecture \cite{vaswani2017}, initially transformative in Natural Language Processing (NLP), spurred the development of Vision Transformers (ViTs) \cite{dosovitskiy2020}. ViTs employ self-attention mechanisms to capture global dependencies across image patches, overcoming the locality bias of CNNs. This led to hybrid models like TransUNet \cite{chen2021} and fully transformer-based architectures like Swin-UNet \cite{cao2022}, which demonstrated improved performance on various medical segmentation tasks by integrating global context. This trend culminated in the emergence of large-scale, pre-trained Foundation Models, exemplified by the Segment Anything Model (SAM) \cite{kirillov2023}. Trained on an unprecedented dataset of over 1 billion masks from 11 million natural images, SAM exhibits remarkable zero-shot generalization capabilities, able to segment arbitrary objects in diverse scenes using simple prompts.
	
	Despite this extraordinary potential, the direct application of SAM to specialized domains like 3D medical imaging is significantly hampered by several critical obstacles. Firstly, SAM's pre-training on natural images creates a substantial \textbf{domain shift} when applied to medical images, which possess distinct characteristics in terms of texture, contrast, noise profiles, and anatomical structure complexity. This often leads to suboptimal performance, particularly in delineating fine details or ambiguous boundaries \cite{ma2024}. Secondly, SAM typically employs a massive ViT encoder (e.g., ViT-H with over 600 million parameters), leading to high \textbf{computational demands}. Fine-tuning such a model requires immense computational resources (high-end GPUs, extensive memory) and large-scale annotated medical datasets, both often unavailable in typical clinical or research settings. Thirdly, the standard SAM architecture has an \textbf{inherent 2D nature}, processing images slice-by-slice and fundamentally ignoring the crucial inter-slice contextual information present in 3D volumetric data. This limits its ability to ensure spatial consistency and accurately segment structures spanning multiple slices.
	
	Concurrently, driven by the need for more efficient sequence modeling, State Space Models (SSMs) have resurfaced as a powerful alternative to Transformers. Originating from control theory, SSMs model systems through latent state representations. A continuous linear SSM is defined by state $h(t) \in \mathbb{R}^N$ and input $x(t) \in \mathbb{R}^D$, governed by:
	\begin{equation} \label{eq:ssm_cont}
		h'(t) = A h(t) + B x(t), \quad y(t) = C h(t) + D x(t)
	\end{equation}
	where $A \in \mathbb{R}^{N \times N}$ is the state matrix, and $B \in \mathbb{R}^{N \times D}$, $C \in \mathbb{R}^{M \times N}$, $D \in \mathbb{R}^{M \times D}$ are projection matrices. For sequence modeling, these are discretized using a timescale parameter $\Delta$, often via the bilinear method, yielding discrete updates:
	\begin{equation} \label{eq:ssm_disc}
		h_k = \bar{A} h_{k-1} + \bar{B} x_k, \quad y_k = \bar{C} h_k + \bar{D} x_k
	\end{equation}
	where $\bar{A}, \bar{B}, \bar{C}, \bar{D}$ are derived from $A, B, C, D$ and $\Delta$. Crucially, this formulation allows modeling long-range dependencies efficiently. Modern SSM architectures, particularly Mamba \cite{gu2023}, introduce a \textbf{selective scan mechanism (S6)} where the matrices $\bar{A}, \bar{B}, \bar{C}$ become input-dependent ($A_k, B_k, C_k$) and incorporate hardware-aware parallel algorithms. This selection mechanism allows the model to dynamically emphasize or forget information based on the input context, while maintaining \textbf{linear time and memory complexity ($O(N)$)} with respect to sequence length $N$. This efficiency advantage over the quadratic complexity ($O(N^2)$) of standard Transformer self-attention is paramount for processing high-resolution medical images or long volumetric sequences where $N$ becomes very large. Visual Mamba (VMamba) \cite{liu2024} further adapted this efficiency to 2D vision tasks using a cross-scan mechanism.
	
	In this paper, we propose to bridge the gap between the powerful generalization of SAM and the efficiency of Mamba by introducing \textbf{Mamba-SAM}, a novel hybrid architectural framework designed for efficient and accurate 3D medical image segmentation. Our central hypothesis is that by keeping the vast knowledge encoded within SAM's pre-trained encoder frozen and integrating it intelligently with lightweight, trainable Mamba-based modules, we can achieve high performance while drastically reducing the computational burden of adaptation. We investigate two primary parameter-efficient strategies embodying this philosophy. The first is a \textbf{Dual-Branch MambaSAM}, which explicitly separates generalist and specialist roles. A frozen SAM ViT encoder acts as the generalist feature extractor, while a parallel, fully trainable VMamba encoder learns domain-specific medical features, fused selectively using a Cross-Branch Attention (CBA) module. The second strategy involves \textbf{3D Adapter MambaSAM}, leveraging Parameter-Efficient Fine-Tuning (PEFT) principles by injecting very lightweight, trainable Tri-Plane Mamba (TP-Mamba) adapters \cite{wang2024} directly into the layers of the frozen SAM ViT encoder to infer 3D volumetric context. Within this adapter framework, we particularly emphasize the \textbf{Multi-Frequency Gated Convolution (MFGC)} variant, which enhances local feature extraction by incorporating frequency domain analysis via 3D DCT and adaptive gating.
	
	This article is a revised and expanded version of a paper entitled ``Mamba-SAM: A Hybrid Architecture for Efficient Cardiac MRI Medical Image Segmentation,'' which was presented at the 16th International Conference on Information and Knowledge Technology (IKT 2025), Iran~\cite{gholipour2025}. Building substantially upon our prior conference work, we now present a comprehensive framework for efficient 3D medical image segmentation. Our expanded contribution introduces a novel parameter-efficient adapter-based paradigm that injects lightweight Tri-Plane Mamba (TP-Mamba) modules directly into SAM's frozen ViT encoder to model true volumetric context—overcoming the inherent 2D slice-wise limitation of the original dual-branch approach. We further propose the Multi-Frequency Gated Convolution (MFGC) module, which enhances feature representation through joint spatial-frequency analysis. Evaluated on the ACDC cardiac MRI dataset~\cite{bernard2018}, our dual-branch \texttt{MambaSAM-Base} model achieves competitive performance (0.906 mean Dice, comparable to UNet++'s 0.907) with state-of-the-art results on the challenging Myocardium (0.910 Dice) and Left Ventricle (0.971 Dice) classes. Meanwhile, the adapter-based \texttt{TP\_MFGC} variant delivers significantly faster inference (4.77~FPS) while maintaining strong accuracy (0.880 Dice), demonstrating that hybridizing foundation models with efficient SSM architectures provides a practical pathway toward clinically viable 3D segmentation systems.
	
	\section{Related Work}
	The evolution of medical image segmentation has been marked by significant architectural shifts, driven by the desire to capture increasingly complex spatial relationships within image data.
	
	\subsection{CNN-based Segmentation: The U-Net Era}
	The U-Net \cite{ronneberger2015} revolutionized biomedical image segmentation with its elegant symmetric encoder-decoder design incorporating skip connections. This allowed for the combination of high-level semantic features with low-level spatial details, proving highly effective for precise localization. Its success spawned a multitude of variations tailored for specific challenges, including the extension to 3D U-Net \cite{cciccek2016} for volumetric data, Attention U-Net \cite{oktay2018} for focusing on salient regions, Residual U-Net \cite{alom2019} for deeper networks, and U-Net++ \cite{zhou2018} with enhanced skip connections. Representing a significant advancement, nnU-Net \cite{isensee2021} provided a self-configuring framework that automatically adapted U-Net architectures and hyperparameters, achieving state-of-the-art results across numerous benchmarks. Despite their successes, CNNs fundamentally rely on local convolution operations, inherently limiting their ability to model long-range dependencies efficiently.
	
	\subsection{Transformers in Medical Vision}
	Inspired by their success in NLP, Transformers entered the medical vision domain to address the limitations of CNNs in capturing global context. ViT \cite{dosovitskiy2020} demonstrated that a pure Transformer could excel at image classification by treating image patches as sequences. Hybrid models like TransUNet \cite{chen2021} combined CNN encoders with Transformer bottlenecks to model both local and global features, showing significant improvements. Architectures such as Swin-UNet \cite{cao2022} leveraged the more efficient Swin Transformer \cite{liu2021swin}, which uses shifted windows for self-attention, enabling hierarchical feature extraction with better scalability. While powerful, the quadratic computational complexity of self-attention mechanisms can still be demanding for high-resolution 3D medical volumes.
	
	\subsection{Foundation Models and SAM Adaptation}
	The concept of foundation models – large models pre-trained on vast datasets capable of adapting to various downstream tasks – gained traction with models like CLIP \cite{radford2021} and culminated in vision with SAM \cite{kirillov2023}. SAM's zero-shot segmentation ability was groundbreaking, but adaptation for medical imaging proved necessary due to the domain gap. Early approaches like MedSAM \cite{ma2024} involved computationally expensive fine-tuning on large medical datasets. More recent strategies focus on Parameter-Efficient Fine-Tuning (PEFT), training only a small fraction of parameters using techniques like adapter modules \cite{houlsby2019} or Low-Rank Adaptation (LoRA) \cite{hu2021lora}. Other works like SAM-Med3D \cite{wang2024sammed3d} introduced 3D-specific modifications. Our approach distinctively keeps the SAM encoder frozen and utilizes Mamba-based PEFT for both domain adaptation and 3D context modeling.
	
	\subsection{State Space Models (Mamba) in Vision}
	SSMs (Equations \ref{eq:ssm_cont}-\ref{eq:ssm_disc}) offer an alternative paradigm for sequence modeling. Mamba \cite{gu2023} revitalized this area with its selective state mechanism (making $\bar{A}, \bar{B}, \bar{C}$ input-dependent) and efficient hardware-aware implementation, achieving linear-time complexity ($O(N)$). This efficiency is a major advantage over Transformers ($O(N^2)$), particularly for the long sequences or high pixel counts common in medical imaging. Its visual variant, VMamba \cite{liu2024}, adapted this efficiency to 2D vision tasks using a cross-scan mechanism, showing performance competitive with ViTs. Mamba's efficiency quickly attracted interest in medical imaging, leading to models like U-Mamba \cite{ma2024umamba}, Mamba-UNet \cite{wang2024b}, VM-UNet \cite{ruan2024}, and SegMamba \cite{xing2024} replacing components in U-Net structures \cite{liao2024, wang2024lkm}. Concurrent work also began exploring SAM-Mamba combinations, often for specific tasks like polyp segmentation \cite{dutta2025}. Our work expands on this by investigating broader architectural strategies (dual-branch vs. adapter) specifically targeting efficient 3D adaptation.
	
	\section{Materials and Methods}
	Our overarching goal is to efficiently adapt the pre-trained SAM model for the specific demands of 3D medical image segmentation, namely handling the domain shift and incorporating volumetric context, while minimizing the number of trainable parameters. We propose two distinct architectural frameworks based on integrating Mamba SSMs.
	
	\subsection{Architecture 1: Dual-Branch MambaSAM}
	This architecture follows a parallel processing strategy, explicitly separating the roles of general feature extraction and domain-specific feature learning. It comprises several key components working in concert.
	
	The \textbf{Frozen SAM Encoder} acts as the generalist. We utilize the standard ViT-B image encoder from SAM, pre-trained on SA-1B, keeping all its parameters ($\approx$ 90M) frozen. For a 2D input slice $X \in \mathbb{R}^{H \times W \times C}$ (resized to $1024\times1024$), it produces a feature map $F_{sam} \in \mathbb{R}^{\frac{H}{16} \times \frac{W}{16} \times D_{sam}}$ ($D_{sam}=768$, possibly projected down). This branch provides robust, high-level semantic understanding.
	
	Operating in parallel is the \textbf{Trainable VMamba Encoder}, acting as the specialist. A smaller VMamba configuration, leveraging Mamba's linear complexity and selective state updates (Eq. \ref{eq:ssm_disc} with input-dependent matrices), is initialized and fully trainable. Processing the same slice $X$, it outputs a feature map $F_{mamba} \in \mathbb{R}^{\frac{H}{16} \times \frac{W}{16} \times D_{mamba}}$ (e.g., $D_{mamba}=384$). This branch learns the low-level features and fine-grained details specific to the medical domain.
	
	Effective merging of knowledge occurs via \textbf{Cross-Branch Attention (CBA) Fusion}. We use a standard multi-head cross-attention mechanism where the specialist VMamba features $F_{mamba}$ serve as the Query (Q), while the generalist SAM features $F_{sam}$ provide the Key (K) and Value (V). This allows domain-specific details to guide the retrieval of relevant general knowledge. The attention output $F_{cba}$ is computed as $\text{softmax}\left(\frac{(F_{mamba}W_q)(F_{sam}W_k)^T}{\sqrt{d_k}}\right)(F_{sam}W_v)$, using trainable projection matrices $W_q, W_k, W_v$.
	
	Finally, a \textbf{Residual Connection and Decoder} combine the features. The CBA output is added residually to the original SAM features ($F_{fused} = F_{sam} + F_{cba}$). This enriched feature map is fed into a decoder (either a simple CNN Decoder using transposed convolutions or an Implicit Feature Alignment (IFA) Decoder \cite{hu2022} using an MLP on continuous coordinates) to generate the final segmentation mask $M \in \{0, 1\}^{H \times W \times N_{classes}}$.
	
	The primary advantage of this architecture is the explicit separation of concerns, allowing dedicated learning of specialized features while benefiting from Mamba's efficiency. Its main limitation remains its inherent 2D nature.
	
	\begin{figure}[htbp]
		\centering
		\includegraphics[width=0.9\textwidth]{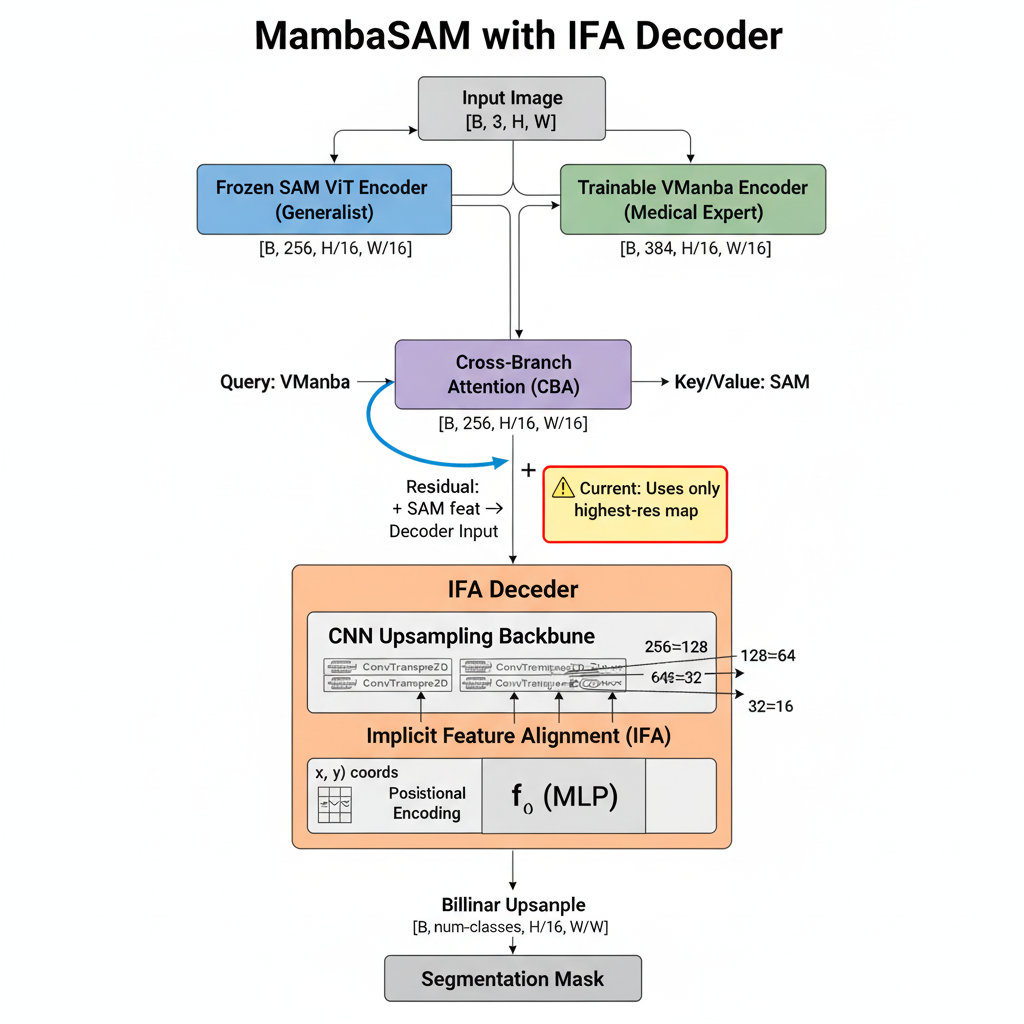}
		\caption{Detailed schematic of the Dual-Branch MambaSAM, highlighting the flow of information, feature dimensions, and trainable components.}
		\label{fig:dual_branch_detailed}
	\end{figure}
	
	\subsection{Architecture 2: 3D Adapter TP-Mamba-SAM}
	This architecture adopts a PEFT approach by modifying the frozen SAM encoder internally with lightweight, 3D-aware Mamba adapters, leveraging Mamba's linear complexity for efficient 3D context modeling.
	
	The core \textbf{Frozen SAM Encoder} (ViT-B) remains unchanged. We then insert \textbf{Trainable TP-Mamba Adapters} \cite{wang2024} after each MSA and MLP block within the ViT encoder, operating residually. If $F_{in}$ is the output from a ViT block, the adapter computes $F_{adapter} = \text{TP-Mamba}(F_{in})$, and the output becomes $F_{out} = F_{in} + F_{adapter}$.
	
	The \textbf{TP-Mamba Adapter Mechanism} involves several steps. First, input 2D feature tokens $F_{in}$ are projected to a lower dimension $D_{adapter}$ and reshaped into a pseudo-3D volume $V$. This volume undergoes parallel feature extraction. A \textbf{Local 3D Context} path captures local spatial patterns. A \textbf{Global 3D Context} path views the volume from three orthogonal planes (Axial HW, Coronal DH, Sagittal DW). Each plane's slices are flattened into sequences and processed by Mamba blocks (using Eq. \ref{eq:ssm_disc} with selective updates) to model long-range dependencies efficiently along each axis. The features from the local and global paths are then fused, projected back to the original SAM dimension $D_{sam}$, and reshaped into token format $F_{adapter}$.
	
	As an enhancement, \textbf{Optional LoRA} \cite{hu2021lora} can be applied to the query, key, and value projections within the frozen MSA layers, allowing highly efficient fine-tuning of the attention mechanism itself.
	
	A significant contribution and variant explored in our work is the \textbf{MFGC Adapter (\texttt{TP\_MFGC})}. Inspired by \cite{xu2025}, we replace the standard multi-scale 3D CNN path within the TP-Mamba adapter with a Multi-Frequency Gated Convolution (MFGC) block. This advanced block enhances the understanding of local features by analyzing them in both spatial and frequency domains, which is particularly beneficial for complex medical textures. Specifically, for an input feature tensor $X_{i}^{s} \in \mathbb{R}^{C_{s} \times D_{s} \times H_{s} \times W_{s}}$ (where $s$ denotes the scale/level), the 3D Discrete Cosine Transform (DCT) is applied. The DCT basis function is defined as:
	\begin{equation} \label{eq:dct_basis}
		D_{d,h,w}^{z_{k},u_{k},v_{k}} = \cos\left(\frac{\pi}{D_{s}}(z_{k}+\frac{1}{2})d\right) \cdot \cos\left(\frac{\pi}{H_{s}}(u_{k}+\frac{1}{2})h\right) \cdot \cos\left(\frac{\pi}{W_{s}}(v_{k}+\frac{1}{2})w\right)
	\end{equation}
	where $(z_k, u_k, v_k)$ are frequency indices and $(d, h, w)$ are spatial voxel coordinates. The input tensor is transformed into the frequency domain using the basis functions:
	\begin{equation} \label{eq:dct_transform}
		X_{i}^{s,k} = \sum_{d=0}^{D_{s}-1}\sum_{h=0}^{H_{s}-1}\sum_{w=0}^{W_{s}-1}(X_{i}^{s})_{:,d,h,w} D_{d,h,w}^{z_{k},u_{k},v_{k}} \in \mathbb{R}^{C_{s}}
	\end{equation}
	This allows the model to explicitly separate low-frequency components (representing overall structure) from high-frequency components (representing fine details, edges, and textures). Following the frequency transformation, channel-wise statistics (average, max, min pooling over frequencies $k$) $Z_{avg}^{s}, Z_{max}^{s}, Z_{min}^{s}$ are computed:
	\begin{equation} \label{eq:freq_pooling}
		Z_{pool}^{s} = \frac{1}{K} \sum_{k=1}^{K} Z_{pool}^{s,k}, \quad pool \in \{avg, max, min\}
	\end{equation}
	These statistics provide a compact summary of the frequency distribution per channel. They then guide a \textbf{Gating Mechanism} (a simple channel attention module with trainable weights $W_1, W_r$ and activation $\delta$, typically ReLU or GELU) that adaptively selects the most relevant frequency components for the segmentation task by computing attention scores $M_i^s$:
	\begin{equation} \label{eq:mfcc_gating}
		M_{i}^{s} = \sigma\left(\sum_{pool \in \{avg, max, min\}} W_{r}(\delta(W_{1}Z_{pool}^{s}))\right) \in \mathbb{R}^{C_{s}}
	\end{equation}
	where $\sigma$ is the sigmoid function. These attention scores $M_i^s$ are then used to modulate the original frequency components $X_i^{s,k}$, effectively enhancing informative frequencies and suppressing noise or irrelevant components. The gated frequency features are transformed back to the spatial domain via Inverse DCT (IDCT), generating a richer and cleaner local feature representation compared to standard convolutions. This MFGC output then replaces the output of the dilated convolution path before fusion with the global Mamba path within the adapter.
	
	Finally, since the adapters introduce 3D awareness, a dedicated \textbf{3D Decoder} using 3D transposed convolutions processes features aggregated from the final encoder blocks to produce the 3D segmentation volume $V_{seg}$.
	
	This adapter-based approach, especially the \texttt{TP\_MFGC} variant utilizing frequency analysis, is highly parameter-efficient and integrates sophisticated 3D context modeling directly within the SAM feature extraction pipeline.
	
	\begin{figure}[htbp]
		\centering
		\includegraphics[width=0.9\textwidth]{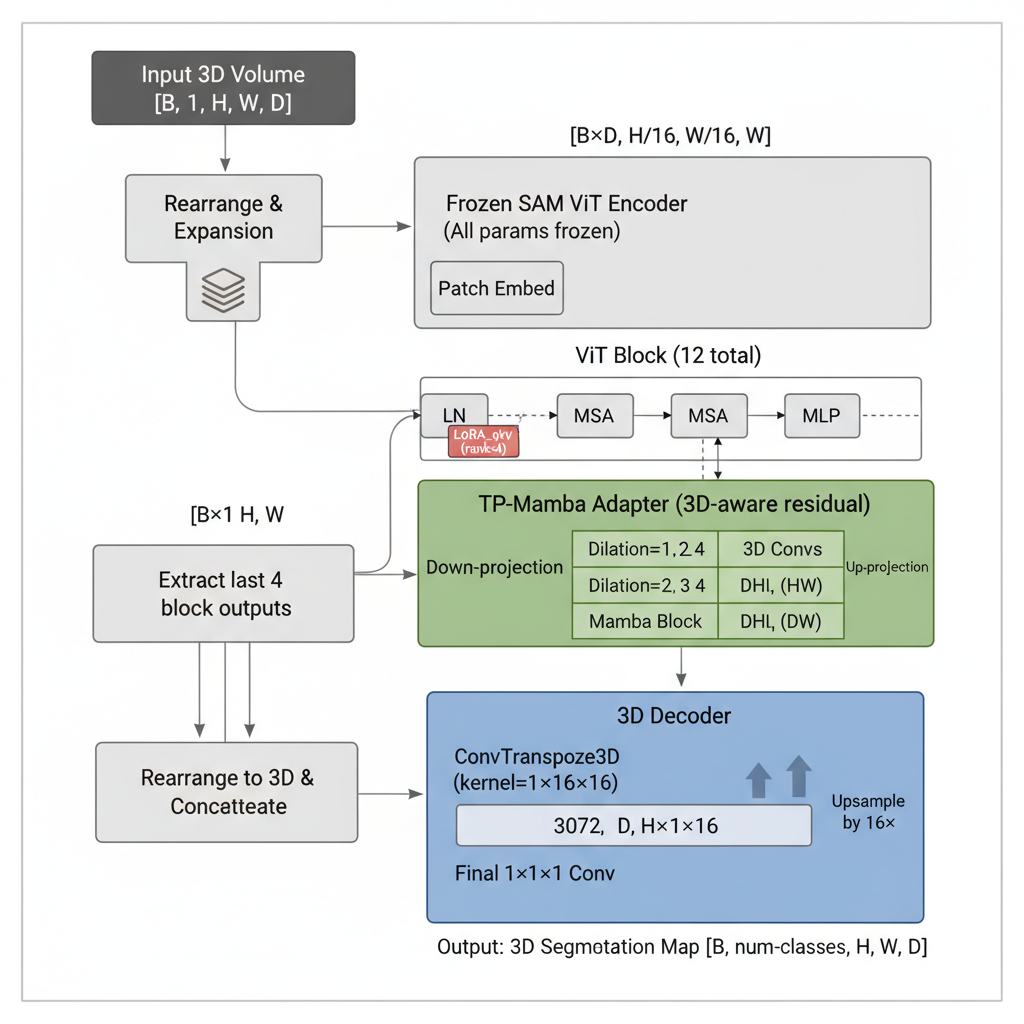}
		\caption{Detailed illustration of how the TP-Mamba adapter (potentially enhanced with MFGC replacing the CNN path) is inserted into a frozen SAM ViT block and its internal mechanism for processing 3D context.}
		\label{fig:tp_mamba_adapter_detailed}
	\end{figure}
	
	\subsection{Dataset and Preprocessing}
	We utilized the ACDC 2017 challenge dataset \cite{bernard2018}, containing 150 cardiac MRI scans with annotations for RV, Myo, and LV at ED and ES phases across 5 patient groups. Our preprocessing pipeline, implemented using MONAI, involved loading NIfTI files, ensuring channel-first format, reorienting to RAS, resampling to isotropic $1.5 \times 1.5 \times 1.5$ mm spacing, normalizing intensities (scaling 0.5-99.5 percentiles to [0, 1]), cropping the foreground, and extracting random patches ($96\times96\times16$ for 3D, $96\times96$ slices for 2D) ensuring label presence. The data was split patient-wise into 70\% training, 15\% validation, and 15\% testing sets.
	
	\subsection{Training Details}
	Training was conducted using PyTorch, MONAI, and Albumentations on Google Colab Pro+ GPUs (H100, A100, L4). We employed the AdamW optimizer with an initial learning rate of $1-2 \times 10^{-4}$, decayed using cosine annealing after a linear warm-up phase. The loss function combined Dice and Cross-Entropy losses ($\mathcal{L}_{total} = \mathcal{L}_{Dice} + \mathcal{L}_{CE}$). Batch sizes varied based on GPU memory. Automatic Mixed Precision (AMP) and gradient clipping (max norm 1.0) were used for stability and efficiency. Model checkpoints were saved based on the best validation Dice score over 50-100+ epochs.
	
	\section{Results}
	We evaluated the proposed Mamba-SAM architectures and compared them against relevant baselines using the ACDC dataset.
	
	\subsection{Evaluation Metrics}
	Standard metrics for semantic segmentation were employed: Dice Similarity Coefficient (DSC) and Intersection over Union (IoU) to measure volume overlap (higher is better), and the 95th Percentile Hausdorff Distance (HD95) in mm to assess boundary accuracy (lower is better). Metrics were computed per class (RV, Myo, LV) and averaged.
	
	\subsection{Quantitative Results Analysis}
	Table \ref{tab:main_results_expanded} presents the detailed quantitative comparison on the ACDC test set. Our dual-branch \texttt{MambaSAM-Base (no neck)} model emerges as the top performer among our proposed architectures in terms of overall Dice score, achieving a mean of 0.906. This result is highly competitive, placing it virtually on par with the heavily optimized UNet++ baseline (0.907) and notably surpassing other methods like Attention UNet (0.894) and SwinUNet (0.654). This strongly suggests the effectiveness of fusing frozen SAM features with a trainable Mamba specialist using cross-attention.
	
	Examining class-specific strengths reveals the exceptional performance of \texttt{MambaSAM-Base (no neck)} on the Myocardium (Myo) and Left Ventricle (LV) classes, where it achieved Dice scores of 0.910 and 0.971, respectively. These scores exceed all other compared models, including UNet++, indicating that this hybrid approach particularly excels at capturing the specific features of these cardiac structures. This might stem from leveraging SAM's general shape priors combined with Mamba's capacity to model fine textures learned during domain-specific training.
	
	The 3D adapter models exhibited varied performance. The \texttt{TP\_MFGC} variant achieved a strong mean Dice score of 0.880, confirming the benefit of injecting 3D context via adapters, particularly with the enhanced MFGC module capturing frequency information crucial for medical image textures. However, its HD95 scores were notably poor, suggesting potential difficulties with fine boundary localization despite good overall volumetric overlap, a point needing further investigation. The simpler \texttt{TP-Mamba} adapter initially performed poorly (0.679 Dice), but incorporating LoRA (\texttt{TP-Mamba (LoRA)}) significantly boosted performance (0.796 Dice) and achieved the best HD95 score for the RV class (4.83 mm). This underscores the importance of PEFT techniques like LoRA for effectively leveraging adapter-based approaches with frozen backbones.
	
	Comparing the two decoder options for the dual-branch model, the simpler CNN decoder (\texttt{MambaSAM-Base (no neck)}) yielded slightly better overall Dice than the IFA decoder (\texttt{MambaSAM (IFA Dec)}, 0.906 vs 0.893). While IFA aims theoretically for sharper boundaries, the standard transposed convolutions performed better in this setup, perhaps requiring less complex tuning.
	
	Regarding boundary accuracy (HD95), UNet++ demonstrated the best overall mean score (2.88 mm). Among our models, \texttt{TP-Mamba (LoRA)} achieved the best RV boundary accuracy (4.83 mm). Our highest-Dice model, \texttt{MambaSAM-Base}, showed moderate HD95 scores (mean 7.53 mm), suggesting a possible trade-off between maximizing volumetric overlap and achieving sub-millimeter boundary precision in this 2D-processing architecture.
	
	\begin{table}[htbp]
		\centering
		\caption{Expanded quantitative comparison on ACDC dataset. Best performance in \textbf{bold}, second best in \textit{italicized}. Our models indicated by \texttt{texttt}.}
		\label{tab:main_results_expanded}
		\resizebox{\textwidth}{!}{%
			\begin{tabular}{@{}lcccccccc@{}}
				\toprule
				& \multicolumn{4}{c}{\textbf{Dice Score (Higher is Better)}} & \multicolumn{4}{c}{\textbf{HD95 (mm) (Lower is Better)}} \\
				\cmidrule(r){2-5} \cmidrule(l){6-9}
				\textbf{Model} & \textbf{RV} & \textbf{Myo} & \textbf{LV} & \textbf{Mean} & \textbf{RV} & \textbf{Myo} & \textbf{LV} & \textbf{Mean} \\
				\midrule
				UNet++ \cite{zhou2018} & \textit{0.898} & 0.871 & 0.952 & \textit{0.907} & 5.01 & 1.14 & 2.50 & \textbf{2.88} \\
				Attention UNet \cite{oktay2018} & 0.878 & 0.858 & 0.947 & 0.894 & 9.05 & 1.44 & \textit{1.23} & 3.91 \\
				AutoSAM (CNN) \cite{liang2025} & 0.888 & 0.860 & 0.942 & 0.897 & 8.68 & 5.48 & 5.63 & 6.60 \\
				MambaUNet \cite{wang2024b} & 0.835 & 0.789 & 0.918 & 0.847 & 7.73 & 2.29 & 3.71 & 4.58 \\
				SwinUNet \cite{cao2022} & 0.572 & 0.607 & 0.782 & 0.654 & 34.46 & 7.38 & 9.71 & 17.18 \\
				\midrule
				\texttt{MambaSAM-Base } \cite{gholipour2025}  & 0.836 & \textbf{0.910} & \textbf{0.971} & \textbf{0.906} & 11.81 & 5.76 & 5.01 & 7.53 \\
				\texttt{MambaSAM (IFA Dec)} & 0.871 & \textit{0.874} & 0.934 & 0.893 & 12.48 & 8.36 & 7.22 & 9.35 \\
				\texttt{TP\_MFGC} & 0.868 & 0.680 & 0.897 & 0.880 & 53.08 & 19.87 & 24.21 & 32.39 \\
				\texttt{TP-Mamba (LoRA)} & 0.758 & 0.769 & 0.860 & 0.796 & \textbf{4.83} & 13.77 & 7.02 & 8.54 \\
				\texttt{TP-Mamba (simple)} & 0.628 & 0.664 & 0.746 & 0.679 & 14.79 & \textit{5.94} & -- & -- \\
				\bottomrule
			\end{tabular}%
		}
	\end{table}
	
	\subsection{Performance, Efficiency, and Resource Usage}
	Table \ref{tab:performance_expanded} details the practical trade-offs in terms of speed, memory, and parameter efficiency. The \texttt{TP\_MFGC} adapter model stands out for its inference speed, processing nearly 5 volumes per second, significantly faster than the slice-based dual-branch models ($\approx$ 2.7 FPS). This efficiency gain stems directly from Mamba's linear complexity ($O(N)$) allowing the 3D adapters to process volumetric context within a single, efficient forward pass, compared to the quadratic complexity of attention if Transformers were used similarly, and the inherently sequential processing of the 2D dual-branch approach.
	
	In terms of memory efficiency, \texttt{TP-Mamba (LoRA)} is exceptional, requiring only 1.90 GB VRAM. This extremely low footprint makes it highly suitable for deployment on resource-constrained devices, albeit with some sacrifice in accuracy. The dual-branch models and \texttt{TP\_MFGC} consume substantially more memory (11-13 GB), typical for larger architectures but still manageable on standard research GPUs.
	
	Regarding parameter efficiency, the dual-branch models and \texttt{TP\_MFGC} train a similar number of parameters ($\approx$ 23-24 M), roughly 20-21\% of their total size, confirming the efficiency compared to full fine-tuning ($\approx$ 90M+ params for SAM encoder alone). \texttt{TP-Mamba (LoRA)} trains the fewest parameters (only adapters and LoRA matrices), highlighting its extreme efficiency. In summary, \texttt{MambaSAM-Base (no neck)} offers the highest accuracy, \texttt{TP\_MFGC} provides the best speed thanks to Mamba's efficiency in 3D processing, and \texttt{TP-Mamba (LoRA)} excels in memory efficiency.
	
	\begin{table}[htbp]
		\centering
		\caption{Expanded performance and resource comparison.}
		\label{tab:performance_expanded}
		\resizebox{\textwidth}{!}{%
			\begin{tabular}{@{}lccccc@{}}
				\toprule
				\textbf{Model} & \textbf{Mean Dice} & \textbf{Inference Speed (FPS/Volume)} & \textbf{Max VRAM (GB)} & \textbf{Total Params (M)} & \textbf{Trainable Params (M)} \\
				\midrule
				\texttt{MambaSAM-Base (no neck)} & \textbf{0.906} & 2.78 & 11.57 & 113.55 & 23.88 ($\approx$ 21\%) \\
				\texttt{MambaSAM (IFA)} & 0.893 & 2.63 & 10.99 & 113.64 & 23.97 ($\approx$ 21\%) \\
				\texttt{TP\_MFGC} & 0.880 & \textbf{4.77} & 12.62 & 112.61 & 22.93 ($\approx$ 20\%) \\
				\texttt{TP-Mamba (LoRA)} & 0.796 & 0.72 & \textbf{1.90} &  $\approx$ 90M + Adapters + LoRA & Very Few (Adapters + LoRA only) \\
				\texttt{TP-Mamba (simple)} & 0.679 & 0.76 & 2.41 & 199.21 & 109.53 ($\approx$ 55\%) \\
				\bottomrule
			\end{tabular}
		}
	\end{table}
	
	\subsection{Qualitative Results Visualization}
\begin{figure}[htbp]
		\centering
		\includegraphics[width=0.9\textwidth]{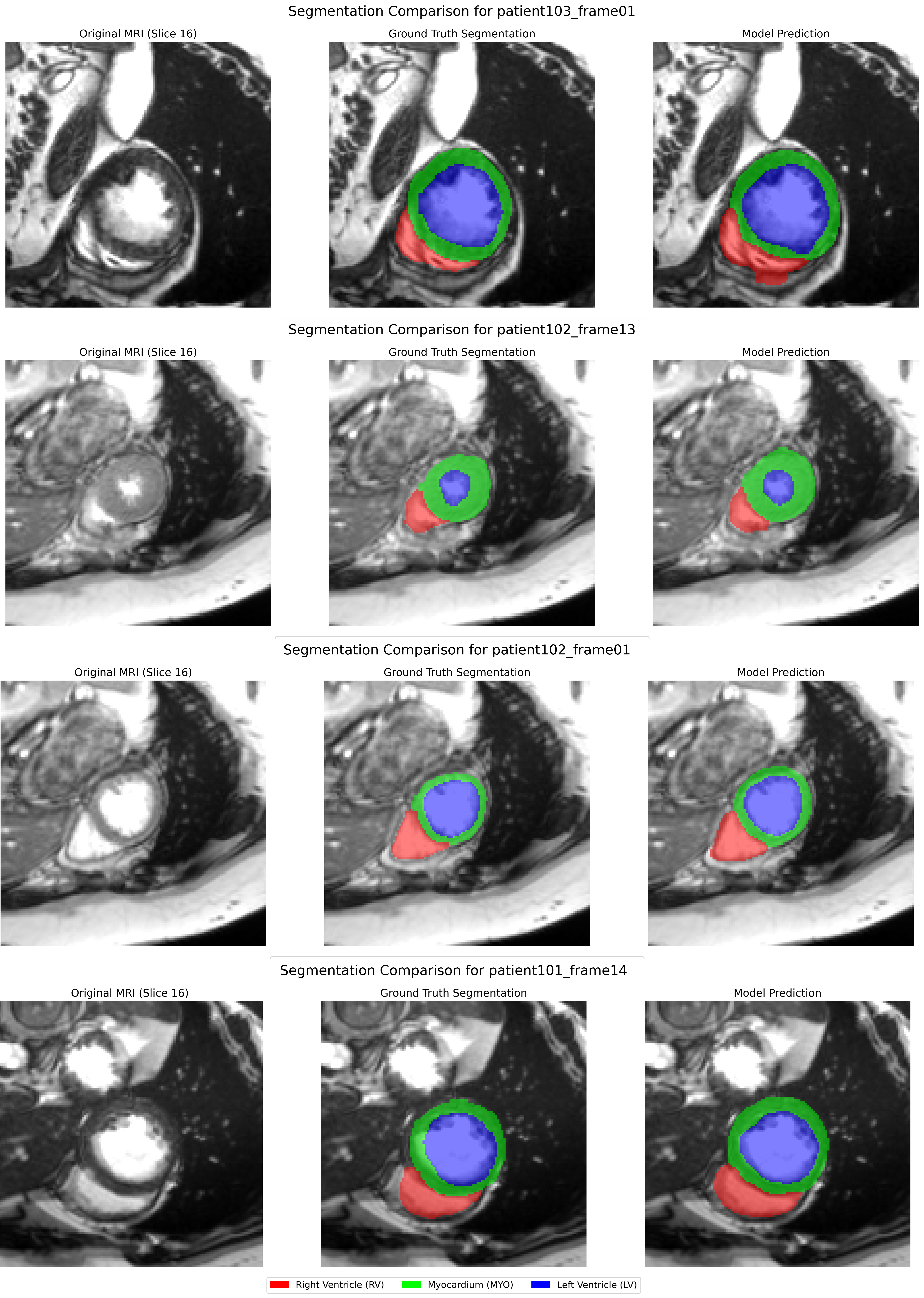}
		\caption{Qualitative segmentation results for the \texttt{TP\_MFGC} model on two different test slices. Red: RV, Green: Myo, Blue: LV. The predictions demonstrate high fidelity to the ground truth boundaries, potentially aided by the MFGC component's frequency analysis.}
		\label{fig:qualitative_detailed}
	\end{figure}

	Visual inspection provides complementary insights. Figure \ref{fig:qualitative_detailed} showcases representative examples from the \texttt{TP\_MFGC} model. The predictions generally align well with the ground truth, accurately capturing the morphology of the RV, Myo, and LV across different patients and slice locations. The MFGC module's ability to consider frequency information may contribute to the good delineation of structures even with subtle contrast variations. While boundary adherence appears good visually, the high quantitative HD95 suggests occasional localized errors might occur, perhaps at complex junctions or where frequency characteristics are ambiguous. In some instances, discrepancies might even relate to potential inconsistencies in the ground truth labels themselves.
	
	\section{Discussion}
	Our investigation into hybrid Mamba-SAM architectures yields several key insights regarding the interplay between accuracy, efficiency, and architectural design. The strong performance of \texttt{MambaSAM-Base}, rivaling UNet++, confirms that combining frozen foundation model features with domain-specific features learned by an efficient trainable architecture like Mamba is highly effective. The cross-attention mechanism appears crucial for enabling the specialist Mamba branch to leverage the generalist SAM branch's knowledge, leading to SOTA results on specific classes like Myo and LV. Mamba's linear complexity makes the specialist branch computationally feasible even alongside the large SAM encoder.
	
	Comparing the dual-branch strategy (\texttt{MambaSAM-Base}) with the adapter-based approach (\texttt{TP-Mamba} variants) reveals a fundamental trade-off. The dual-branch model provided superior accuracy, suggesting that a separate, fully trainable Mamba encoder allows more comprehensive learning of complex domain-specific features than lightweight adapters modifying existing SAM features. However, the adapter-based models, particularly \texttt{TP\_MFGC}, offered significantly better inference speed. This advantage stems from the ability of Mamba-based adapters to process 3D context efficiently (linear complexity) within a single forward pass through the modified encoder, contrasting with the slice-by-slice processing of the dual-branch model. The enhancement provided by the MFGC module over the simpler CNN path in the adapter highlights the benefit of incorporating advanced feature analysis techniques like frequency domain processing, which is particularly relevant for medical images with complex textures. The extreme memory efficiency of \texttt{TP-Mamba (LoRA)} further highlights the benefits of adapter-based PEFT for low-resource deployment. The choice depends on the application's priorities: accuracy favors the dual-branch, while speed and memory efficiency favor adapters, with MFGC offering a strong balance between speed and accuracy within the adapter paradigm.
	
	The importance of incorporating 3D context, even through lightweight adapters like in \texttt{TP\_MFGC}, is evident from its strong Dice score and efficiency gains over the 2D dual-branch model. However, the poor performance of the simple \texttt{TP-Mamba} indicates that careful adapter design (e.g., MFGC leveraging frequency information) or complementary techniques (like LoRA) are necessary for effectiveness. The surprisingly poor HD95 of \texttt{TP\_MFGC} suggests a potential weakness in boundary localization despite good volumetric overlap, possibly due to the MFGC focusing more on texture than precise edges, warranting further analysis.
	
	A key advantage of our main approaches is their parameter efficiency. By training only a fraction ($\leq$ 25M) of the total parameters, we validate the feasibility of adapting large foundation models like SAM without the prohibitive cost of full fine-tuning, making SOTA techniques more accessible. Mamba's inherent parameter efficiency contributes significantly here compared to potentially using Transformer-based adapters.
	
	Our study faced limitations due to computational constraints, preventing exhaustive training of all variants (like memory-augmented or distilled models) and evaluation on multiple datasets. Based on our findings, future work should focus on optimizing hybrid architectures, completing the knowledge distillation pipeline for a potentially highly efficient student model, developing multi-modal text-guided segmentation capabilities, evaluating cross-dataset generalization, and investigating the HD95 anomaly observed with \texttt{TP\_MFGC}.
	
	\section{Conclusions}
	This paper introduced Mamba-SAM, a novel hybrid architectural framework designed to efficiently adapt the Segment Anything Model (SAM) for the demanding task of 3D medical image segmentation. By strategically combining the powerful, frozen general-purpose features of SAM with the linear-time efficiency and long-range modeling capabilities of Mamba-based State Space Models, we successfully addressed the critical challenges of domain shift, 2D architectural limitations, and prohibitive computational costs associated with fine-tuning large foundation models.
	
	We presented and evaluated two primary parameter-efficient adaptation strategies: a dual-branch architecture explicitly fusing features via cross-attention, and an adapter-based approach injecting 3D-aware Mamba modules, notably enhanced with Multi-Frequency Gated Convolution (MFGC), into the frozen SAM encoder. Our results on the ACDC cardiac MRI dataset demonstrate the efficacy of this hybrid paradigm. The dual-branch \texttt{MambaSAM-Base} model achieved a mean Dice score of 0.906, performing on par with highly optimized state-of-the-art methods like UNet++ while setting new benchmarks for Myocardium and Left Ventricle segmentation accuracy. Concurrently, our adapter-based \texttt{TP\_MFGC} model showcased superior inference speed due to Mamba's efficiency in handling 3D context, coupled with strong accuracy, and the \texttt{TP-Mamba (LoRA)} variant highlighted potential for deployment in extremely memory-constrained environments.
	
	Our findings validate that hybridizing frozen foundation models with trainable, efficient architectures like Mamba offers a compelling balance of high accuracy, computational efficiency (leveraging $O(N)$ complexity), and parameter efficiency. This work paves the way for broader adoption of powerful foundation models in resource-aware clinical and research settings, representing a practical and potent direction for the future of automated medical image analysis.

	\section*{Funding}
	This research received no external funding.
	
	\section*{Data Availability Statement}
	The ACDC dataset used in this study is publicly available from the Medical Segmentation Decathlon challenge website: \url{https://acdc.creatis.insa-lyon.fr/}.
	
	\section*{Acknowledgments}
	The authors would like to thank the creators of the ACDC dataset for making it publicly available. The computational resources were provided by Google Colab Pro+ platform.
	
	\section*{Conflicts of Interest}
	The authors declare no conflicts of interest.
	
	\bibliography{references}
	
\end{document}